\documentclass[conference]{IEEEtran}
\IEEEoverridecommandlockouts
\usepackage{amsmath,amssymb,amsfonts}
\DeclareMathOperator{\E}{\mathbb{E}}
\usepackage{multirow}
\usepackage{algpseudocode}
\usepackage{algorithm}
\usepackage{graphicx}
\usepackage{textcomp}
\usepackage{xcolor}
\usepackage{hyperref}
\usepackage[sorting = none]{biblatex}
\def\BibTeX{{\rm B\kern-.05em{\sc i\kern-.025em b}\kern-.08em
    T\kern-.1667em\lower.7ex\hbox{E}\kern-.125emX}}
\begin{document}
\title{Planning with RL and episodic-memory behavioral priors}

\author{\IEEEauthorblockN{Shivansh Beohar}
\IEEEauthorblockA{\textit{IIIT Allahabad} \\
Prayagraj, India \\
shivansh.bhr@gmail.com}
\and
\IEEEauthorblockN{Andrew Melnik}
\IEEEauthorblockA{\textit{University of Bielefeld} \\
Bielefeld, Germany \\
andrew.melnik.papers@gmail.com}
}

\maketitle

\begin{figure}[htbp]
\centerline{\includegraphics[scale=0.358]{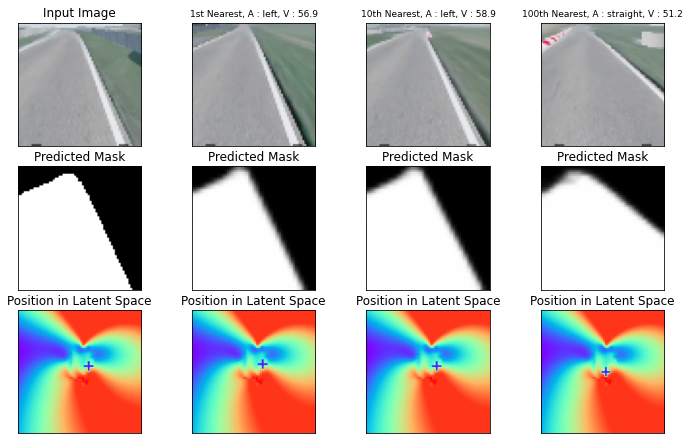}}
\caption{
Comparison of states at different distances from a query state. The first column represents the query state. The second, third, and forth columns represent the states at the 1st, 10th and 100th positions in the sorted list of states by their distance (see eq. \ref{distance equation}) from the query state. A - the action and V - the value of the state. The first row shows the observation, the second row shows the semantic mask, the third row shows the position in the 2D latent space and its value indicated by color (red - high value, blue - low value state).}
\label{planner_3row}
\end{figure}

\begin{abstract}
The practical application of learning agents requires sample efficient and interpretable algorithms. Learning from behavioral priors is a promising way to bootstrap agents with a better-than-random exploration policy or a safe-guard against the pitfalls of early learning. Existing solutions for imitation learning require a large number of expert demonstrations and rely on hard-to-interpret learning methods like Deep Q-learning. In this work we present a planning-based approach that can use these behavioral priors for effective exploration and learning in a reinforcement learning environment, and we demonstrate that curated exploration policies in the form of behavioral priors can help an agent learn faster.
\end{abstract}


\begin{figure}[htbp]
\centerline{\includegraphics[scale=0.20]{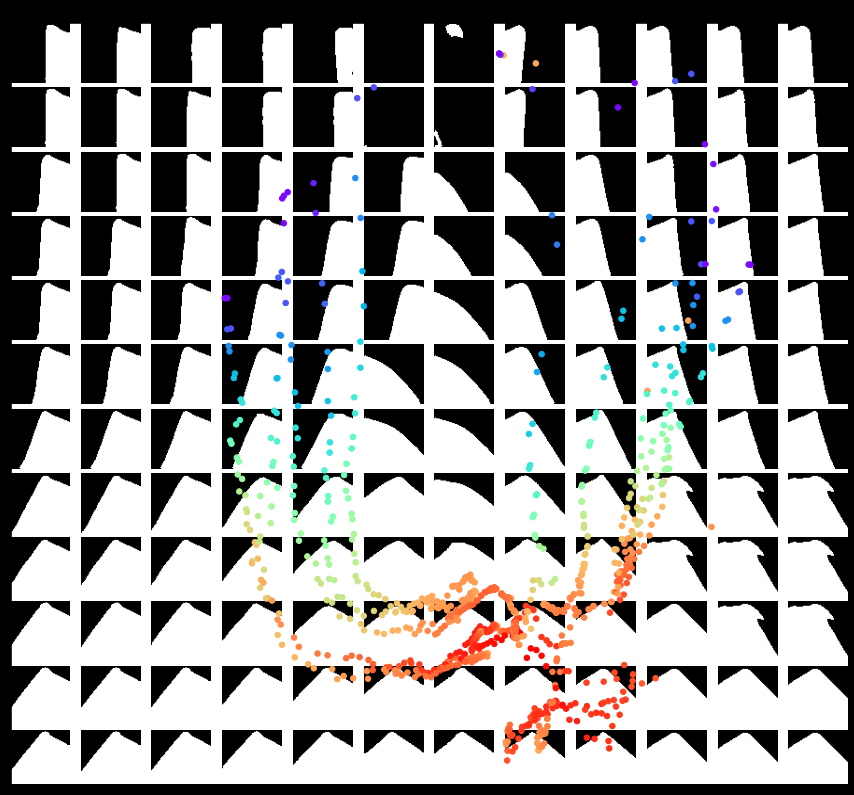}}
\caption{Collage of semantic masks overlayed with points in latent space. The scattered points are colored according to their values, red - high value and blue - low value.}
\label{latent_grid}
\end{figure}

\section{Introduction}
Reinforcement learning (RL) agents often struggle from cold start problems, especially in the case of long horizon tasks \cite{bach2020learn}\cite{konen2019biologically}\cite{schilling2018approach}\cite{harter2020solving}. It arises due to random initialization of models leading to poor exploration and thus very little learning in initial phase of training.

To solve this problem, imitation learning \cite{10.1145/3054912}\cite{torabi2019recent} algorithms introduce behavior priors. Expert demonstrations in the form of $SARS'$ tuples are provided to the agent prior to phase-III (RL) training. The agents thus starts with a knowledge of “good” or “safe” actions to take in most of the situations, leading to more effective exploration in the phase-III (RL) and hence better sample efficiency. As the agent explores, it gathers the missing knowledge for states not encountered in demonstrations, and eventually learns skills at par (and beyond) with the expert. A recent popular example of this approach is DQfD \cite{hester2018deep}. Existing Q-Learning-based methods require large amounts of data to create robustness to low-level pixel noise and high-level feature noise. This coupled with dynamically changing environment \cite{bach2020error} make it complicated to use such networks for practical application. Moreover, the learned policies are not interpretable, leading to difficulty in validating such models.

In order to make learning more robust, discrete and interpretable \cite{melnik2019modularization} we demonstrate our Planner module. It is loosely based on the Monte Carlo tree search \cite{browne2012survey}, which evaluates different alternatives from the current state, conditioned on different actions to a certain depth, and choose the best on the basis of expected reward.

\section{Exploration with behavioral priors}

The learning starts with using an initial controller $C$ that executes the action given by behavior priors. Essentially, the goal is to gather tuple sequences $(S, A, R, S', D)$ of variable length $T$ returned by the environment, in order to learn value of states, given the initial controller $C$, and initialize planning component which leads to more efficient exploration with priors in further phases of learning. For our experiments we focus on Learn-to-Race (L2R) environment \cite{herman2021learntorace}\cite{L2R}\cite{beohar2022solving}, the details of the environment are given in Section \ref{Experiment section}. In L2R, we modelled the behavior prior as changing direction when an \textit{unsafe} state is encountered. To determine if the state is \textit{safe}/\textit{unsafe}, we let the agent drive straight at a constant velocity until the episode ends. Due to curved tracks, this means that driving straight will always end in a negative reward due the car moving out of the track. For these episodes, we mark the last \textit{unsafe\_offset} frames as \textit{unsafe} and the remaining as \textit{safe} states.
The agent is trained in phased manner, starting from behavior prior, followed by RL phase (Phase-III). In behavior prior based phase-I the agent was trained to determine low value states as described above (also see Appendix, Algorithm \ref{Straight drives}), and in phase-II it explores actions in these low valued states (Algorithm \ref{Unsafe state exploration algorithm}). To improve further and fill in the missing knowledge, we use planner (See Appendix, Algorithm \ref{Planner Algorithm for Selecting Action}) to train the agent in RL phase (phase-III) as shown in Algorithm \ref{RL Phase Training}.   

\section{Planning with episodic memory behavioral priors}

\subsection{State representation}
\label{state representation section}
We trained U-Net \cite{ronneberger2015u} and (Sigma) VAE \cite{rybkin2020sigmavae} models on the collected pictures from the previous phases to translate visual representations into the latent space of the VAE. For encoding the states, instead of using RGB images returned from the environment, the algorithm encodes into latent dimensions the semantic masks of the road. The selection of the right semantic is important for performance \cite{melnik2021critic} and is a design choice for different environments. For L2R, we selected the masks of road-track as the semantic information. However, other environments may require different encoding schemes.

Mathematically,
\begin{equation}
   E(O) = V(U(O))
\end{equation}

where $E$ is the encoding of the observation $O$. $U$ is semantic extractor U-Net ($F_{Unet}$) model. $V$ is a sigma variational autoencoder ($F_{VAE}$) to encode the semantic masks extracted by U-Net. The parameters of these models are provided in Appendix.

\subsection{Episodic Memory}
The agent estimates an expected reward value of states using trajectories collected with initial behavioral prior policy $C$. We use low valued states as the primary candidates to explore actions, which leads to targeted exploration in the initial learning phase.

Episodic memory $M$ consisting of $N$ experienced trajectories, where $S_a^b$ is state at time step $a$ in the $b^{th}$ trajectory, $A_a^b$ is action, $R_a^b$ is observed reward and $I_a^b$ is any other auxiliary information are collected in the following form, each tuple is called a Trajectory Point (TP) $
[ (S_1^1,A_1^1,R_1^1,I_1^1), (S_2^1,A_2^1,R_2^1 ,I_2^1), \cdots
(S_{t1}^1,A_{t1}^1,R_{t1}^1,I_{t1}^1)$ , $(S_1^2,A_1^2,R_1^2,I_1^2) \cdots  (S_{tn}^N,A_{tn}^N,R_{tn}^N,I_{tn}^N)] $ Once stored, the planner $P$ selects the best trajectory point $TP_{best}$ from the episodic memory $M$ based on criteria listed below and executes action taken from that trajectory for a given query state i.e $ P(M,S) = TP_{best} : \E (R_i | S_i , A_i )$ is maximized over all i.

Apart from state-action-reward we also require other information like a pointer to the next TP in a trajectory and other convenience attributes like Grid coordinates (see Latent Grid), pointer to k-th step TP in future, step index and trajectory ID etc. for various update and retrieval tasks. For all algorithmic purposes, consider TP as a data structure which stores all the above values.

\subsection{Latent Grid and Storing Trajectories}
To reduce noise, artifacts and dimensionality of input observation space, we encode the observations into low dimensional space of dimensionality $= m$ (for L2R, $m = 2$) using the method described in section \ref{state representation section}. Since we aim to leverage searchability of states in this lower dimensional latent space, we quantize the latent into a grid $L$ of size $g$ (for L2R, it will be a $2$ dimensional grid of size $g$ = $100$x$100$). Each cell in the grid represents a range of continuous latent encoding depending on the range of values in the stored DB. Mathematically, to convert an encoding ($E$ : $R^m$) into latent grid coordinates (G : $N^m$ ) of matrix with size = $g$, we use the formula ($E$ is collection of all encoding value) $ G_i = floor( E_i / ( max(E) - min(E) ) + g//2)  \forall$ i in $[1 \cdots m]$

Effectively, each grid cell stores all a list of all TPs that lie on it, and stores their average value as the its cell value for future use.
\begin{figure}[htbp]
\centerline{\includegraphics[scale=0.27]{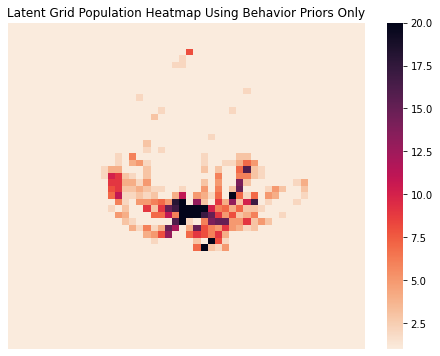} \includegraphics[scale=0.27]{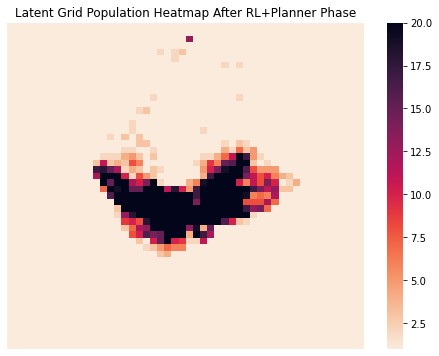} }
\caption{100x100 Latent Grid population heatmap before and after Training. The latent space organization changes to some extent as the VAE is retrained.}
\label{popden}
\end{figure}

As the agent explores the environment and visits each cell, the cell accumulates the value $VC$ as the average of the sum of discounted future rewards for each TP lying on grid coordinate $C$. Fig. \ref{popden} demonstrates how the cells are populated by explorations during training phases.

\subsection{Retrieving Trajectory Points.}

In order to select the action to be taken, the planner takes as an input the last $p$ states observed and the episodic memory stored in the latent grid and ranks the neighboring stored trajectory points based on an estimate of best state values in their near future of $k$ time steps. 

The ideology of above design is as follows:
\begin{itemize}

\item Searching with the correct context : The Markov state model assumption that the future state only depends on the current is often violated in a practical application environment. We aim to alleviate some of this limitation by selecting that trajectory which is similar to the current at least for the past $p$ states. The distance between two state encodings $s_1^0$ and $s_2^0$ , both $m$ dimensional, therefore is :

\begin{multline}
\label{distance equation}
    \Delta (s^1 , s^2 ) =   \sum_p ( \| s^1_{-p}  - s^2_{-p} \|^2 + \| s^1_{-p+1}  - s^2_{-p+1} \|^2  \\ + \cdots   \| s^1_0 - s^2_0 \|^2 )
\end{multline}

    where $s_n$ is state occurring $n$ timesteps before $s_0$ and $\|. \|^2$ is the L2 norm.
    
This ensures that there is more correlation between the matches than mere visual similarity. This is important in environments where context plays a significant role and is not captured by state information alone.

\item Searching in a limited area : While searching for a match in the latent space, we only consider points which are very close to query state. This avoids matches with high value states which do not corresponds to the query state, and gives an indication to the planner to explore rather than exploit.
We consider TPs which lie within $n$ steps of 8-connected grid cells as close neighbors, for L2R we have set $n = 1$. 

\item Criteria of selection of ideal TP :  For those states in immediate vicinity of current state, we only consider the value of those state in next $k$ time steps. This is ensure that following these points leads to high rewards.
\end{itemize}

The planner effectively explores when the best match is not a good values state, otherwise explores an action which is not yet explored in the region, a complete algorithm is given in Appendix Algorithm \ref{Planner Algorithm for Selecting Action}

\begin{algorithm}

 \caption{Unsafe state exploration}
 \label{Unsafe state exploration algorithm}
 \begin{algorithmic}[1]
 \renewcommand{\algorithmicrequire}{\textbf{Input:}}
 \renewcommand{\algorithmicensure}{\textbf{Output:}}
 \Require 
 No. of episodes to explore \textit{e} (= 20), Controller \textit{C}, Environment \textit{env}, Action Space \textit{A}, Phase-I Database \textit{DB}, Phase-II Database \textit{DB2 ( = $\phi$ )}, Gamma : \textit{g}, U-Net : \textit{$F_{Unet}$}, VAE : \textit{$F_{VAE}$}

 \Ensure Trajectories of explored actions in \textit{unsafe} states
 \For{i $=$ 0 to e-1}
 \State done $\longleftarrow$ false
\State step $\longleftarrow$  0
\State ep\_data $\longleftarrow$ {}
\State unsafe $\longleftarrow$  False
  \While {not done} 
  \State step += 1
\State Sample action a $\longleftarrow$  C(A,unsafe)
 \Comment{If Unsafe returns a random action from A, else take the default action. For L2R default is moving forward with constant velocity}
\State obs, reward, done $\longleftarrow$  env.step(a)
\State unsafe $\longleftarrow$ is\_unsafe( obs,DB,$F_{Unet}$ , $F_{VAE}$)

\State ep\_data $\longleftarrow$  ep\_data $\bigcup$ {[obs,reward,i,step,done]}
\EndWhile

\State ep\_data $\longleftarrow$  augment(ep\_data,g)
\Comment{Compute and store future discounted rewards}
\State DB2 $\longleftarrow$ DB2 $\bigcup$ ep\_data
\EndFor \\
\Return $DB2$ 
 \end{algorithmic} 
 \end{algorithm}

\begin{algorithm}[t]

 \caption{Phase-III-RL Training}
 \label{RL Phase Training}
 \begin{algorithmic}[1]
 \renewcommand{\algorithmicrequire}{\textbf{Input:}}
 \renewcommand{\algorithmicensure}{\textbf{Output:}}
 \Require 
 Latent Grid \textit{L} , No. of training episode \textit{n\_train\_ep}, No. of previous state to match trajectories \textit{p}, Maximum distance to qualify as a neighbor \textit{n} , Maximum neighbors to select for sorting \textit{q}, Value function lookahead range \textit{k} , U-Net \textit{$F_{UNet}$} ,  VAE \textit{$F_{VAE}$} , Phase-II Database \textit{DB}, Environment \textit{env}
\Ensure Trained Latent Grid
\For { i $=$ 0 to n\_train\_ep}
\State E $\longleftarrow$ Deque( MAX\_LEN $=$ p )             
\Comment{ [$E_{-p} ,  E_{-p+1} , .... E_0$]} 
 \State    obs $\longleftarrow$ env.reset()

 \For {i $=$ 0 to p}:
 \Comment{initialize last p deque with starting frame}
 \State       E $\longleftarrow$ E $\bigcup$ ($F_{VAE}$( $F_{UNet}$( obs ) ))
 \State   best = None
\While{ not done }
\State g\_nn $\longleftarrow$ neighbors(L , E, n, q)
\Comment{Select close neighbors in vicinity}
\State  best $\longleftarrow$ sort(ngrid\_max(g\_nn, k))
\Comment{Sort TPs in g\_nn by the maximum value estimate in max(t, t+k) timesteps (k$=$10)}
 \If{length(best) $=$ 0}
 \Comment{If the value of best neighbor is not $>$ min(10$\%$), then take the least explored action.}
 \State             action $\longleftarrow$ Random()
\ElsIf {best[0].ngrid\_max $<$ min(10$\%$)}
 \State             best $\longleftarrow$ sort(ngrid\_max(g\_nn, k))
 \State             action $\longleftarrow$ Least\_explored(best)
\Else
 \State             action $\longleftarrow$ best[0].action
\EndIf
 \State           obs, reward, done $\longleftarrow$ env.step(action)
  \State          e $\longleftarrow$ $F_{VAE}$( $F_{UNet}$( obs ) )
  \State          E $\longleftarrow$ E $\bigcup$ (e)
   \State         ep\_data $\leftarrow$ ep\_data $\cup$ [ obs, reward, i, step, done ]
\EndWhile
  \State    ep\_data $\longleftarrow$ augment(ep\_data,g)
  \Comment{Compute and store future discounted rewards}

   \State   DB $\longleftarrow$ DB $\bigcup$ ep\_data
 \State     Train\_VAE(DB)
 \Comment{Finetune VAE on DB}

  \State    Update(DB, L, $F_{VAE}$)
  \Comment{Update models}
\EndFor

\EndFor \\
\Return L 

 \end{algorithmic} 
 \end{algorithm}

\section{Experiments}
\label{Experiment section}
\begin{figure}[htbp]
\centerline{\includegraphics[scale=0.27]{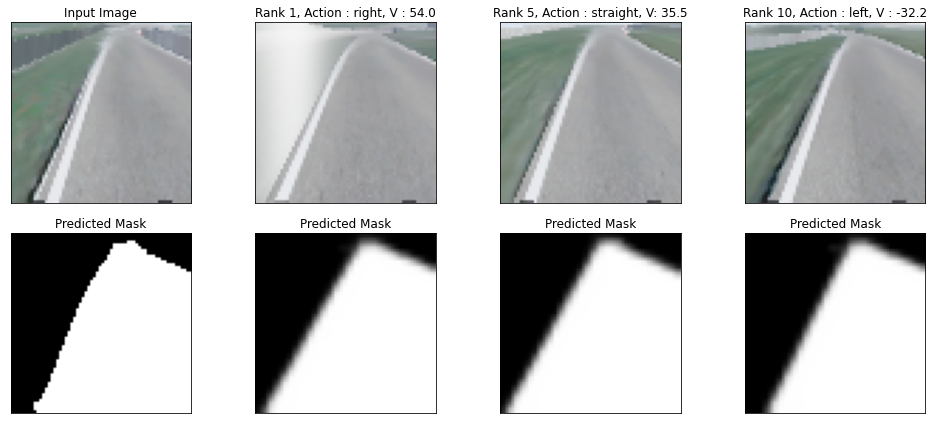}}
\caption{Comparison of nearest states with different executed actions. V - the value of the state. The states in the immediate vicinity of the query states are ranked according their value. The value validate the choice to rank the alternatives.}
\label{best_matches_planner}
\end{figure}

To demonstrate the applicability of our approach we deploy our agent on L2R environment, the environment is \textit{Formula 1} style single player racing simulator, which provides first person view of the car as shown in the Fig. \ref{planner_3row} and Fig. \ref{best_matches_planner} along with velocity information. The action space have two actions Steering and Acceleration, both of which can be between $[-1 \dots 1]$. It features realistic physics and visuals. The reward at every time step is proportional to the percentage of track completed in that time step and a penalty which is a large negative award associated with going out of the track with at least two wheels outside the road area.
Our approach works with continuous action space as well as with discrete action space, something a lot of other approaches cannot provide. For the discrete action space, we have restricted steering exploration to only three values, that is $-1$ (turn right), $0$ (drive straight) and $1$ (turn left). 
We trained the U-Net and VAE model using the observations collected in Phase-I, with VAE finetuning in Phase-III (RL) as well.

\begin{figure}[htbp]
\centerline{\includegraphics[width=0.485\columnwidth]{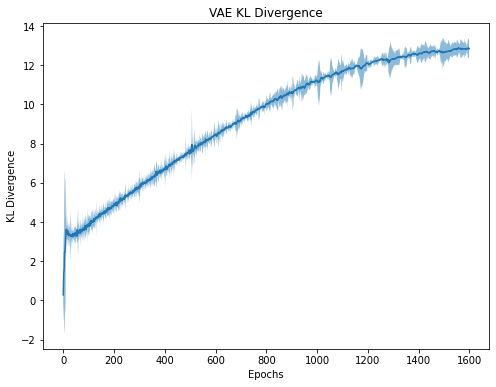}
\includegraphics[width=0.515\columnwidth]{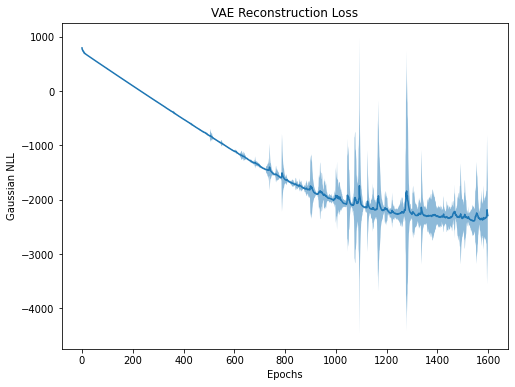}}
\caption{Training Loss curves for VAE}
\label{vae_training}
\end{figure}

See Fig. \ref{vae_training} for training curves for VAE.

\begin{figure}[htbp]
\centerline{\includegraphics[width=\columnwidth]{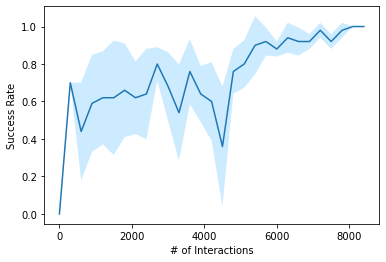}}
\caption{Success Rate graph for RL training phase (Phase-III). Average of 5 agents.}
\label{RL_training}
\end{figure}

The latent space shows good separation between visually distinct scenes as shown in Fig. \ref{latent_grid}. The 
performance metrics and comparisons are shared in Fig \ref{vae_training}, Fig \ref{RL_training} and Table \ref{comp_table}.

We have presented results of various agents which were trained on a seen track in L2R environment, and evaluated for 3 laps on unseen track. We have also included Human (Expert), Random agent and a Model Predictive Controller baselines for comparison. The MPC model follows a center line of the track at moderate speed and is treated as a naive policy for the task. The metrics we have chosen for our study are: success rate, which is an indicator of how much \% of the track an agent could complete without failing and average speed, reported as distance covered per unit time (KM/H). See Appendix, Algorithm \ref{Agent Evaluation} for evaluation algorithm.

\begin{table}
\begin{tabular}{ |p{2.5cm}||p{2.5cm}|p{2cm}| }
 \hline
 Agent& Success Rate & Average Speed (KMPH)\\
 \hline
Human & 100\% ($\pm 0$) & 114.0 ($\pm 2.3$) \\
Random & 1\% ($\pm 0.6$) & 9.4 ($\pm 1.5$) \\
MPC & 69.5\% ($\pm 10.7$) & 40.5 ($\pm 0.1$) \\
RL-SAC & 11.8\%($\pm 0.1$) & 22.1 ($\pm 1.5$) \\
\textbf{RL-Planner (Ours)} & \textbf{93.3\%($\pm 0.04$)} & \textbf{59.3 ($\pm 0.2$)} \\
 \hline
\end{tabular}
\vspace{5pt}
\caption{Agent Performance Comparison. All results except ours is taken from \cite{herman2021learntorace}}
\label{comp_table}
\end{table}

Table 1 shows the comparison with the above baselines, along with another popular deep learning approach Soft Actor Critic. The Soft Actor Critic required ~48 hours of training, compared to ~2 hours of training for our algorithm (incl. behavior priors, training of VAE, U-Net and Phase-III (RL) ) on a NVIDIA GTX 1660Ti GPU, a training time interaction vs performance graph for our algorithm is given in Fig. \ref{RL_training}, the graph is an average of 5 training runs, with each training episode checkpoint evaluated 3 times. The agent was able to achieve 100\% success rate after 24 episodes, which is equivalent to 40K environment interactions.

To demonstrate the performance of planner, we present a planner point of view for a query state. Fig \ref{planner_3row} shows different stored trajectory points sorted according to the distance to the query state as described in Eqn. \ref{distance equation}.
The planner's decision process can be seen in Fig. \ref{planner_3row} depicting the comparison of a state with other states at different distances and Fig. \ref{best_matches_planner} which depicts the closest neighbors of a query state.

\section{Discussion}
We presented a behavior prior initialized agent outperforming SAC on L2R task with substantially less learning interactions with the environment. The planner module utilizes stored episodic memory to learn value estimates and guide the agent towards efficient exploration. The approach enables sample efficient adaptation and a few-shot learning policy \cite{schilling2021decentralized}. We have shown with our results that a planning approach, combined with behavior priors assessments, can enable a sample efficient few-shot learning and provide interpreted agent decisions.

\printbibliography

@article{10.1145/3054912,
author = {Hussein, Ahmed and Gaber, Mohamed Medhat and Elyan, Eyad and Jayne, Chrisina},
title = {Imitation Learning: A Survey of Learning Methods},
year = {2017},
issue_date = {March 2018},
publisher = {Association for Computing Machinery},
address = {New York, NY, USA},
volume = {50},
number = {2},
issn = {0360-0300},
url = {https://doi.org/10.1145/3054912},
doi = {10.1145/3054912},
journal = {ACM Comput. Surv.},
month = {apr},
articleno = {21},
numpages = {35}
}

@article{torabi2019recent,
  title={Recent advances in imitation learning from observation},
  author={Torabi, Faraz and Warnell, Garrett and Stone, Peter},
  journal={arXiv preprint arXiv:1905.13566},
  year={2019}
}

@article{beohar2022solving,
  title={Solving Learn-to-Race Autonomous Racing Challenge by Planning in Latent Space},
  author={Beohar, Shivansh and Heinrich, Fabian and Kala, Rahul and Ritter, Helge and Melnik, Andrew},
  journal={arXiv preprint},
  year={2022}
}

@inproceedings{hester2018deep,
  title={Deep q-learning from demonstrations},
  author={Hester, Todd and Vecerik, Matej and Pietquin, Olivier and Lanctot, Marc and Schaul, Tom and Piot, Bilal and Horgan, Dan and Quan, John and Sendonaris, Andrew and Osband, Ian and others},
  booktitle={Proceedings of the AAAI Conference on Artificial Intelligence},
  volume={32},
  number={1},
  year={2018}
}

@article{browne2012survey,
  title={A survey of monte carlo tree search methods},
  author={Browne, Cameron B and Powley, Edward and Whitehouse, Daniel and Lucas, Simon M and Cowling, Peter I and Rohlfshagen, Philipp and Tavener, Stephen and Perez, Diego and Samothrakis, Spyridon and Colton, Simon},
  journal={IEEE Transactions on Computational Intelligence and AI in games},
  volume={4},
  number={1},
  pages={1--43},
  year={2012},
  publisher={IEEE}
}

@inproceedings{ronneberger2015u,
  title={U-net: Convolutional networks for biomedical image segmentation},
  author={Ronneberger, Olaf and Fischer, Philipp and Brox, Thomas},
  booktitle={International Conference on Medical image computing and computer-assisted intervention},
  pages={234--241},
  year={2015},
  organization={Springer}
}

@misc{rybkin2020sigmavae,
    title={Simple and Effective VAE Training
    with Calibrated Decoders},
    author={Oleh Rybkin and Kostas Daniilidis
    and Sergey Levine},
    year={2020},
}

@inproceedings{bach2020learn,
  title={Learn to move through a combination of policy gradient algorithms: Ddpg, d4pg, and td3},
  author={Bach, Nicolas and Melnik, Andrew and Schilling, Malte and Korthals, Timo and Ritter, Helge},
  booktitle={International Conference on Machine Learning, Optimization, and Data Science},
  pages={631--644},
  year={2020},
  organization={Springer}
}

@inproceedings{konen2019biologically,
  title={Biologically-inspired deep reinforcement learning of modular control for a six-legged robot},
  author={Konen, Kai and Korthals, Timo and Melnik, Andrew and Schilling, Malte},
  booktitle={2019 IEEE international conference on robotics and automation workshop on learning legged locomotion workshop,(ICRA) 2019, Montreal, CA, May 20-25, 2019},
  year={2019}
}

@article{melnik2019modularization,
  title={Modularization of end-to-end learning: Case study in arcade games},
  author={Melnik, Andrew and Fleer, Sascha and Schilling, Malte and Ritter, Helge},
  journal={arXiv preprint arXiv:1901.09895},
  year={2019}
}

@inproceedings{schilling2018approach,
  title={An approach to hierarchical deep reinforcement learning for a decentralized walking control architecture},
  author={Schilling, Malte and Melnik, Andrew},
  booktitle={Biologically Inspired Cognitive Architectures Meeting},
  pages={272--282},
  year={2018},
  organization={Springer}
}

@misc{herman2021learntorace,
      title={Learn-to-Race: A Multimodal Control Environment for Autonomous Racing}, 
      author={James Herman and Jonathan Francis and Siddha Ganju and Bingqing Chen and Anirudh Koul and Abhinav Gupta and Alexey Skabelkin and Ivan Zhukov and Andrey Gostev and Max Kumskoy and Eric Nyberg},
      year={2021},
      eprint={2103.11575},
      archivePrefix={arXiv},
      primaryClass={cs.RO}
}

@inproceedings{bach2020error,
  title={An error-based addressing architecture for dynamic model learning},
  author={Bach, Nicolas and Melnik, Andrew and Rosetto, Federico and Ritter, Helge},
  booktitle={International Conference on Machine Learning, Optimization, and Data Science},
  pages={617--630},
  year={2020},
  organization={Springer}
}

@inproceedings{melnik2021critic,
  title={Critic guided segmentation of rewarding objects in first-person views},
  author={Melnik, Andrew and Harter, Augustin and Limberg, Christian and Rana, Krishan and S{\"u}nderhauf, Niko and Ritter, Helge},
  booktitle={German Conference on Artificial Intelligence (K{\"u}nstliche Intelligenz)},
  pages={338--348},
  year={2021},
  organization={Springer}
}

@article{schilling2021decentralized,
  title={Decentralized control and local information for robust and adaptive decentralized Deep Reinforcement Learning},
  author={Schilling, Malte and Melnik, Andrew and Ohl, Frank W and Ritter, Helge J and Hammer, Barbara},
  journal={Neural Networks},
  volume={144},
  pages={699--725},
  year={2021},
  publisher={Elsevier}
}

@article{harter2020solving,
  title={Solving physics puzzles by reasoning about paths},
  author={Harter, Augustin and Melnik, Andrew and Kumar, Gaurav and Agarwal, Dhruv and Garg, Animesh and Ritter, Helge},
  journal={arXiv preprint arXiv:2011.07357},
  year={2020}
}

@misc{L2R,
title ={{Learn-to-Race Autonomous Racing Virtual Challenge}},
NOTE = "\url{https://www.aicrowd.com/challenges/learn-to-race-autonomous-racing-virtual-challenge
}; accessed May 14, 2022."
}
\onecolumn
\section*{\Huge{Appendix}}

\setcounter{algorithm}{2}
\begin{algorithm}

 \caption{Planner Algorithm for Selecting Action}
 \label{Planner Algorithm for Selecting Action}
 \begin{algorithmic}[1]
 \renewcommand{\algorithmicrequire}{\textbf{Input:}}
 \renewcommand{\algorithmicensure}{\textbf{Output:}}
 \Require 
 Latent Grid \textit{L}, Query state along with p-1 past states \textit{$E_q$}, Maximum distance to qualify as a neighbor \textit{n}, Maximum neighbors to select for sorting \textit{q}, Value function lookahead range \textit{k}, Minimum threshold value of state to be considered as good state to follow \textit{val\_thresh}
\Ensure Action to execute
  \State g\_nn $\longleftarrow$ neighbors(L , $E_q$, n, q)
  \Comment{For all neighbors lying in $\pm$ n grid coordinate as $E_0$ , return top q (q$=$10) neighbors which are closest to E according to the sum of p (p$=$3 history size) element wise L2 distance as described above}
  \State best $\longleftarrow$ sort(ngrid\_max(g\_nn, k))
  \Comment{Sort TPs in g\_nn by the maximum value estimate in max(t, t+k) timesteps (k$=$10)}
  \If{best = $\phi$ OR best[0].ngrid\_max $<$ val\_thresh:}
  \Comment{If the value of best neighbor is not $\>$ min(10$\%$), then increase the search to the entire latent grid. This is guaranteed to yield a positive match, but can be a poor match and slower.}
  \State g\_nn $\longleftarrow$ neighbors(L ,Eq, $\infty$)
  \State  best $\longleftarrow$ sort(ngrid\_max(g\_nn, k))
  \EndIf
  \State action $\longleftarrow$ best[0] action
  \Comment{Copy the action of the best match} \\
 \Return $action$ 
 \end{algorithmic}
 \end{algorithm}

\begin{algorithm}

 \caption{Collection of trajectories with the initial controller $C$}
 \label{Straight drives}
 \begin{algorithmic}[1]
 \renewcommand{\algorithmicrequire}{\textbf{Input:}}
 \renewcommand{\algorithmicensure}{\textbf{Output:}}
 \Require No. of episodes to explore \textit{e} (= 20), Controller \textit{C},
Environment \textit{env}, Database \textit{DB} (= $\phi$), Gamma \textit{g}, Unsafe\_offset \textit{unsafe\_offset}
 \Ensure  Trajectories
  
  \For {$i = 0$ to $e-1$}
  \Comment{LOOP for each episodes}
  \State  done $\longleftarrow$  false \Comment{Explore till the episode ends} 
\State step $\longleftarrow$  0
\State ep\_data $\longleftarrow$  $\phi$
\While{not done}
\State   step += 1
\State  Sample action a $\longleftarrow$  C
\Comment{Query Controller for a naive exploration (behavioral prior)}
\State        obs, reward, done $\longleftarrow$  env.step(a)
\State        ep\_data $\longleftarrow$  ep\_data $\bigcup$ {[obs,reward,i,step,done]}

\EndWhile 

\State    ep\_data $\longleftarrow$ augment(ep\_data,g,unsafe\_offset)
\Comment{Compute and store future discounted rewards, mark last unsafe\_offset frames of each episodes as unsafe}

\State    DB $\longleftarrow$  DB $\bigcup$  ep\_data
\EndFor

 \Return $DB$ 
 \end{algorithmic} 
 \end{algorithm}

\begin{algorithm}[H]

 \caption{Agent Evaluation}
  \label{Agent Evaluation}
 \begin{algorithmic}[1]
 \renewcommand{\algorithmicrequire}{\textbf{Input:}}
 \renewcommand{\algorithmicensure}{\textbf{Output:}}
 \Require 
 Trained latent Grid \textit{L}, No. of previous state to match trajectories \textit{p} , Maximum distance to qualify as a neighbor \textit{n} , Maximum neighbors to select for sorting \textit{q}, Value function lookahead range \textit{k}, U-Net \textit{Un} , VAE \textit{vae}, Environment \textit{env}

\textbf{Initialization}
\State E $\longleftarrow$ Deque( MAX\_LEN $=$ p )             
\Comment{ [$E_{-p} ,  E_{-p+1} , .... E_0$]} 
 \State    obs $\longleftarrow$ env.reset()

 \textbf{Evaluate}

 \For {i $=$ 0 to p}
\State         E.append(Vae( Un( obs ) ))
\EndFor
\Comment{Check if best is one of $D_n$ nearest neighbors of $E_p$}

 \State    best $\longleftarrow$ None

\For {step $=$ 0 to $\infty$ }

 \State          g\_nn $\longleftarrow$ neighbors(L , E, n, q)
 \Comment{For all neighbors lying in $\pm$ n grid coordinate as $E_0$ , return top q (q$=$10) neighbors which are closest to E according to the sum of p (p$=$3 history size) element wise L2 distance}
 \State          best $\longleftarrow$ sort(ngrid\_max(g\_nn, k))
 \Comment{Sort TPs in g\_nn by the maximum value estimate in max(t, t+k) timesteps (k$=$10)}
 \If    {length(best) $=$ 0 OR best[0].ngrid\_max $<$ 0}
 \Comment{If the value of best neighbor is not $>$ min(10$\%$), then increase the search to the entire latent grid. This is guaranteed to yield a positive match, but can be a poor match and slower.}
\State                g\_nn $\longleftarrow$ neighbors(L , [GX,GY], $\infty$)
 \State               best $\longleftarrow$ sort(ngrid\_max(g\_nn, k))

\EndIf
 \State    action $\longleftarrow$ best[0].action
     
 \State     obs,reward $\longleftarrow$ env.take\_action(action)
 \State     e $\longleftarrow$ Vae( Un( obs ) )
 \Comment {Convert observation to encoding}
 \State     E.append(e)
 \State     Store(obs, action, reward, step)
\EndFor

 \end{algorithmic} 
 \end{algorithm}

\subsection{Model Training}

\subsubsection{U-Net}
We collected image masks from the environment during the Phase-I-Straight Driving (around 600 image-mask pairs) and applied Brightness (0.2), Contrast (0.2), Color Jitter (0.2), Hue(0.2), Flip (0.5), Scale (min = 0, max = 0.2) and Rotation (min = 0, max = 0.3) transformation for data augmentation.
The architecture is as described in \cite{ronneberger2015u} with input image size = $64$ x $64$
We trained the model for 60 epochs, with lr = $0.0001$ without any LR scheduling. We used Adam Optimizer for our experiment.
\subsubsection{VAE}
We collected image masks from the environment during the Phase-I training (around 600 images)
The architecture is as described in \cite{rybkin2020sigmavae} with input image size = $28$ x $28$ and $z_dim$ = 2.
We trained the model for 150 epochs, with lr = $0.001$ without any LR scheduling. We used Adam Optimizer for our experiment.

\subsection{Planner}
For Learn2Race \cite{herman2021learntorace} environment we have used the following parameters for our planner :
Past states to consider for matching TP \textit{p} $=3$, Maximum distance to qualify as a neighbor \textit{n} $=1$, Maximum neighbors to select for sorting \textit{q} $=10$, Value function lookahead range \textit{k} $=10$, Minimum threshold value of state to be considered as good state to follow \textit{val\_thresh} $=0$

\end{document}